\documentclass[10pt, conference, compsocconf]{IEEEtran}
\usepackage{cite}

\usepackage[latin1]{inputenc} 
\usepackage[T1]{fontenc}      

 \usepackage[pdftex]{graphicx}
\usepackage[cmex10]{amsmath}
\usepackage{array}

\usepackage{mdwmath}

\usepackage[tight,footnotesize]{subfigure}
\usepackage{url}

\begin{document}
%
\title{Interpreting weight maps in terms of cognitive or clinical neuroscience: nonsense?}




%
\author{\IEEEauthorblockN{J. Schrouff\IEEEauthorrefmark{1}\IEEEauthorrefmark{2} and
J. Mour\~{a}o-Miranda\IEEEauthorrefmark{1}\IEEEauthorrefmark{2}
}
\IEEEauthorblockA{\IEEEauthorrefmark{1}Centre for Medical Image Computing, Department of Computer Science, University College London, UK}
\IEEEauthorblockA{\IEEEauthorrefmark{2}Max Planck University College London Centre for Computational Psychiatry and Ageing Research, University College London, UK}}



\maketitle

\begin{abstract}
Since machine learning models have been applied to neuroimaging data, researchers have drawn conclusions from the derived weight maps. In particular, weight maps of classifiers between two conditions are often described as a proxy for the underlying signal differences between the conditions. Recent studies have however suggested that such weight maps could not reliably recover the source of the neural signals and even led to false positives (FP). In this work, we used semi-simulated data from ElectroCorticoGraphy (ECoG) to investigate how the signal-to-noise ratio and sparsity of the neural signal affect the similarity between signal and weights. We show that not all cases produce FP and that it is unlikely for FP features to have a high weight in most cases.
\end{abstract}


%
\IEEEpeerreviewmaketitle

\section{Introduction}
Linear machine learning models can be seen as providing two outputs: predictions and weight maps. The latter shows the relative contribution of the individual features to the model and has been heavily used in the neuroimaging community to infer conclusions about brain structure/function. There has however been a recent debate on whether weight maps can provide information about the neural signals leading to a significant classification/regression model \cite{Haufe2014,Haynes2015a,Kia2017}. The authors of \cite{Haufe2014} indeed suggest that weight maps provide a poor recovery of the input neural signal and lead to false positives. They further demonstrate that the amplitude of the weight does not reflect the amplitude of the signal difference in a feature. However, their examples are specific cases with low signal-to-noise ratio (SNR). Here, we investigate the recovery of two widespread techniques, namely SVM \cite{Cortes1995} and sparse MKL \cite{Rakotomamonjy2008} when varying the SNR, as well as the distribution of simulated neural signals.

\section{Material}

\subsection{Original data}
The data was recorded from intracranial electrodes implanted in a patient with pharmaco-resistant epilepsy. This study was approved by the Stanford IRB and the patient gave written consent to participate in the study. 64 electrodes were implanted and signal was recorded using a Tucker Davis system (sampling rate: 1536Hz, Fig.\ref{fig1}A). The signal was acquired during a 5-minute wakefulness rest period, with eyes closed. Electrodes displaying pathological activity were discarded from further analysis.

\subsection{Simulated design}
\label{simdes}
A fake experimental design was simulated: 2 conditions, `A' and `B', presented at random every 1.9 seconds. The stimuli are further assumed to last for 1 second. This yielded 146 stimuli, 73 for each category.

\subsection{Pre-processing}
\label{preproc}
Signal pre-processing was performed with specific ECoG routines\footnote{\url{github/LBCN/Preprocessing_Pipeline}} using Matlab\footnote{\url{www.mathworks.com}} and SPM12\footnote{\url{www.fil.ion.ucl.ac.uk/spm}}. First, the data was converted to SPM format and downsampled to 1kHz. The continuous signal was filtered for line noise and harmonics (stop-band: 57-63Hz, 117-123Hz, 177-183Hz) and an automatic quality assessment identified `noisy' or `spiky' channels based on their variance and number of `jumps' (i.e. signal derivative$>100\mu V$), leaving 38 `good' channels ($p_c$). The data was re-referenced to the average of all good channels before being epoched in the $[-400, 1400]$ms window around `onset' (as defined in \ref{simdes}) and baseline corrected using the $[-400, 0]$ms window. Epochs displaying flat segments of more than 4ms or `jumps' larger than $100\mu V$ were discarded from further analysis. The signal was then decomposed using a 5-wavelets decomposition in the 70 to 170Hz frequency band (step: 10Hz, avoiding 120Hz) to estimate High Frequency Broadband (HFB) power. The time-frequency signal was z-scored based on the pooled baselines of all events in the $[-300, 0]$ms window before onset to avoid edge effects. The signal in the $[-200, 1200]$ms window was finally smoothed using a 50ms Gaussian window. Epochs displaying z-scores larger than 8 were discarded, leaving 60 trials for condition `A' and 56 for `B'. Further analysis focused on the $[0, 1000]$ms window after `onset'. This pre-processing procedure is standard in stimulus-based ECoG studies (e.g. \cite{Mesgarani2014}, see Fig.\ref{fig1}B for example signals).

\subsection{Simulation of signals}

\begin{figure*}[!t]
\centering
\includegraphics[width=16cm]{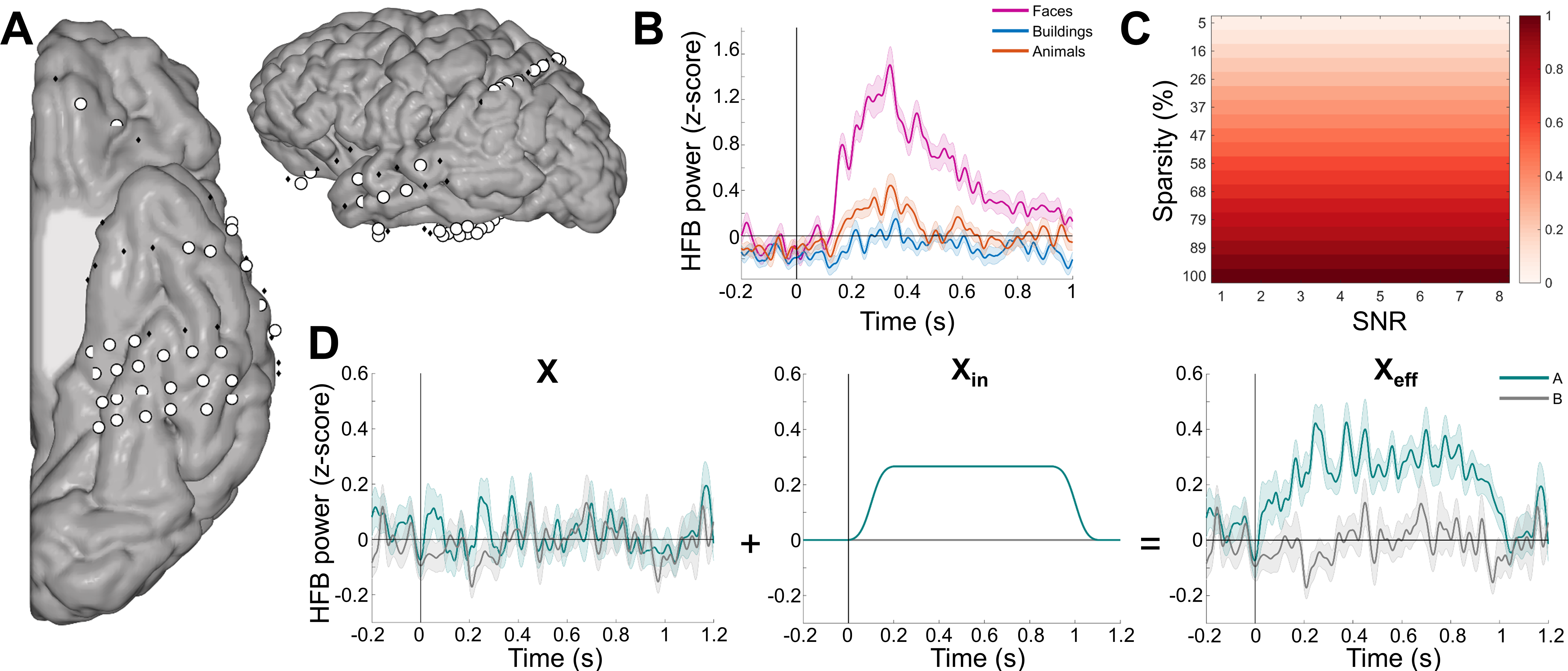}
\caption{\textbf{Data: A} Spatial sampling on the considered subject. White circles represent channels assessed as `good' ($p_c = 38$). \textbf{B} Example of ECoG stimulus-based signals, in z-score, when the same subject is viewing images of human faces (pink), buildings (blue) or animals (orange). Pre-processing is similar to \ref{preproc}. The estimated SNR for human faces is 5.0567, with an average amplitude of 0.5298 in the $[0, 1000]$ms window. \textbf{C} Input sparsity pattern for each combination of sparsity and $SNR_{in}$. For each $SNR_{in}$, a $S_{in}$ proportion of channels were `turned on', from 5$\%$ to 100$\%$, as represented by the color bar. \textbf{D} For each `on' channel, a smoothed rectangular is added to condition A trials to obtain the desired $SNR_{in}$ (=4 in this example).}
\label{fig1}
\end{figure*}

All modifications of data structure were performed on the pre-processed data to avoid an effect of the pre-processing on the obtained results. 
To simulate neural signal, a rectangular window was added to all epochs of condition A in the $[0, 1000]ms$ after onset. For realistic purposes, this window was smoothed by a 200ms Gaussian window. Our approach allowed to vary two aspects of the `input' neural signal: the Signal-to-Noise Ratio (SNR) and the Sparsity ($S$, see Fig.\ref{fig1}C,D).

\begin{itemize}
\item \textit{SNR}: the amplitude of the signal in the rectangular window was computed based on a desired SNR on each channel:

$$X_{A,eff} = X_{A} + SNR_{in} \times std(\overline{X}_{B})$$

Where $X_{A,eff}$ represents the amplitude of the effective simulated signal for condition A trials, $X_{A}$, the amplitude of the raw signal for trials A, $SNR_{in}$, a fixed number representing the desired SNR and $\overline{X}_{B}$, the average trace of B trials (i.e. our `noise'). $SNR_{in}$ was varied from 1 to 8 by steps of 0.5. These numbers were chosen to obtain both non-significant and significant discrimination between the `A' and `B' trials. In real ECoG datasets, the distribution of estimated SNR varies from -4 to 10 (estimated from comparing viewing images of human faces to viewing images of body parts from multiple species, in 8 subjects).

\item \textit{S}: the number of channels on which the SNR was modified varied from 2 to 38 (over 38), by a step of 2 (i.e. from 5.6$\%$ to 100$\%$, with a step of 5.6$\%$). The sparsity of the model is defined here as the number of channels on which a simulated signal is added, over the total number of channels,: $$S_{in}=\frac{|I_{in}|}{p}$$ With $I_{in}$, the set of features with a non-zero input and $p$ the number of features ($38 \times 1001$). This definition corresponds to that of \cite{Baldassarre2017}. Real ECoG datasets are often quite sparse, with typically only a few channels being selective for a certain condition (e.g. $S=2$ to $20\%$, observations from 16 datasets, proportion of channels selective for human faces). 
\end{itemize}

The simulated signal is added on channels, using a contiguous (and long) time window. This choice reflects our observations of ECoG signals and the obtained traces resemble signals from stimulus-based experiments. The order of the channels on which the signal was added was assigned in a pseudo-random fashion. This means that the simulated signal is added independently of the channel spatial position and the only spatial smoothness of the signals derive from the correlated `noise' structure present in the resting-state data.

We hence obtain two stages of input data: the simulated neural input (smoothed rectangular window or 0, $X_{in}$) and the `effective' input data (simulated data plus correlated noise, $X_{eff}$). We are then interested in recovering the time series of the difference between `A' and `B' trials for the input `neural' signals, which here corresponds to $X_{in}$ (as there is no input for `B') . This difference signal was also averaged over time points for each channel, to obtain channel level estimations ($\overline{X}_{in}$). 

The data set and Matlab function to generate the simulated signals are available open-source\footnote{\url{https://github.com/JessicaSchrouff/Simulated_ECoG}}.

\section{Methods}

\subsection{Modelling}
\subsubsection{Univariate testing}
A univariate permutation test assessed the significance of the difference between `A' and `B' epochs. On each channel, it computed the average across time in each trial ($\overline{X}_{A-B,c}=\sum_{t=1}^{p_T} X_{A-B,c}$, $p_T$ being the number of time samples, i.e. 1001) and tested whether the median difference between the 2 conditions was larger than when the conditions were permuted. 5,000 permutations were estimated on each channel. The obtained p-values were corrected for multiple comparisons (FDR correction for the number of channels). The test returns which channels display a significant difference in z-score across the 2 conditions.

\subsubsection{Machine learning}
All machine learning modeling was performed in PRoNTo version 3 \cite{Schrouff2013a,Schrouff2016}. For each channel, a linear kernel was built by selecting the $[0, 1000]$ms time window after onset. Two machines were considered in this example: a SVM \cite{Cortes1995} and a sparse MKL \cite{Rakotomamonjy2008} grouping the features by channel. The kernels were added and then mean centered before SVM classification. Kernels were mean centered and normalized for MKL modelling. The classification between `A' and `B' trials was based on a 10-fold cross-validation scheme (leave-epochs-per-class-out). The soft-margin hyper-parameter of the machine was optimized (grid search: C=0.01, 0.1, 1, 10, 100, 1000) using a nested 5-fold cross-validation. For machine learning based models, the balanced accuracy, averaged across folds, assessed model performance. In addition, 500 permutations assessed the significance of this performance for SVM models (uncorrected $p<0.01$). For each model, a weight map was derived (one weight per feature, i.e. time point on each channel). In addition, the contribution per kernel (i.e. per channel, $W_c$) was returned by the MKL model. To obtain a `channel-wise' contribution for the SVM model, channel averages were computed based on \cite{Schrouff2013}:
$$ W_c = \frac{\sum_{t=1}^{p_T} |w_{c,t}|}{p_T}$$

Based on the channel contributions (either SVM or MKL), an `Expected Ranking' (ER) was defined \cite{Schrouff2013}, with channels ranked in descending order based on their $W_c$ for each fold and the result being averaged across folds (and rounded). The highest possible rank is $p_c$ while channels with no contribution to the model have a rank of 0.

\subsection{Recovery metrics}
\subsubsection{Univariate}
The performance and recovery of the univariate test are implicitly  related. We estimated the True Positive rate (TP) as the proportion of channels that were assessed as significant when `neural signal' was added. Similarly, the False Positive rate (FP) computed the number of channels assessed as significant while no neural signal was added.

\subsubsection{Machine learning}
To assess the recovery of the machine learning models, we compared the obtained weight maps with the simulated difference signal. To this end, we estimated the cosine distance between $W$ and $X_{in}$, at the feature level, and then at the channel level (i.e. $W_c$ and $\overline{X}_{in}$). Finally, we estimated the TP and FP rates for each model as:

$$ TP = \frac{|ER > thresh|}{S_{in}\times p_c}$$
$$ FP = \frac{|ER > thresh|}{thresh}$$

Where the threshold $thresh$ corresponds to (1) the ranking $p_c-(S_{in} \times p_c)$, which is the lowest ranking that can be expected if all channels $\in X_{in}>0$ were ranked correctly at the top of the ranking, and (2) $p_c$ - 10, an arbitrary number that is often selected in publications (i.e. looking at the Top 10). Both metrics vary between 0 and 1. $FP=1$ means that all channels without `neural input' had a higher ranking than the defined threshold (which varies in case (1)).

\section{Results}

\begin{figure*}[!t]
\centering
\includegraphics[width=17cm]{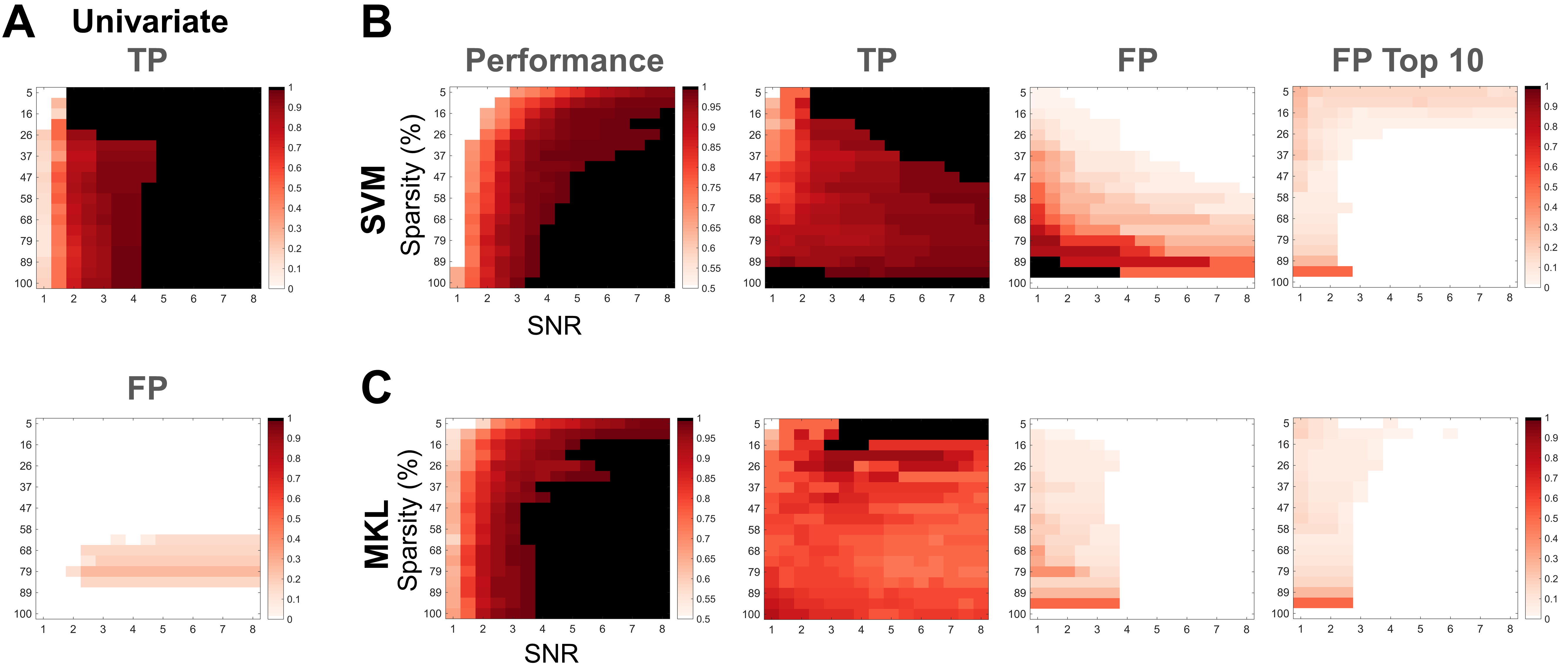}
\caption{\textbf{A} Univariate TP (top) and FP (bottom) rates, color-coded for each SNR-$S$ combination. Black means all TP were correctly identified in the TP plot. White in the FP plot means no FP. \textbf{B} SVM results: Performance (in \%, black means perfect performance while white means no significant results), TP rate, FP rate computed based on a varying ER threshold and FP rate in Top 10. Note that for $S=100\%$, all channels are 'positive', so FP is undefined. \textbf{C} MKL results.}
\label{Fig2}
\end{figure*}

\subsection{Performance}
The performance of the SVM and MKL models are displayed in Fig.\ref{Fig2}. Even for low SNR, both models are significant when the signal is distributed across a minimum of features (non-significant models are displayed in white). SVM performs perfectly in 38$\%$ of the cases while MKL performs perfectly in 50$\%$ of the cases. In addition, MKL outperforms SVM ($>+5\%$ accuracy) in 20$\%$ of the cases.

\subsection{Recovery}
\subsubsection{Univariate}
In only 5 cases, the test cannot identify any significant result (shown in white). Its TP rate is then quite low for low SNR values. In 60$\%$ of the cases, the test correctly identifies all TP (Fig.\ref{Fig2}A). For high SNR ($SNR_{in}>4$), false positives are detected, which suggests that FDR correction is not strict enough in those cases.

\subsubsection{Machine learning}
The average cosine distance between the weight maps and input difference signal (at the feature and channel levels) are presented in Table \ref{table1}\footnote{The patterns are not presented in this publication due to space constrains}. They show an overall good similarity between SVM weights and the difference signal for high SNR and distributed patterns. For MKL, there is a maximum overlap and similarity area between MKL weights and the input signal, covering high SNR and low $S$, as expected from the algorithms' priors. Interestingly, the average similarity between weights and `neural' difference signal increases when considering channel level signals ($\overline{X}_{in}$) and model contributions ($W_c$).

\begin{table}[!h]
\centering
\caption{Cosine distance between weight maps and the input neural difference signal ($X_{in}$), at the feature ($W$) and channel ($\overline{W}_c$) level.}

{\renewcommand{\arraystretch}{1}
\renewcommand{\tabcolsep}{0.17cm}
\newcolumntype{C}{ >{\centering\arraybackslash} X}
\begin{tabular}{|l|c|c||c|c|}
\hline
  & \multicolumn{2}{c|}{\textbf{SVM}} & \multicolumn{2}{c|}{\textbf{MKL}}\\
\hline
\textbf{Signal} & $W$ & $W_c$ & $W$ & $W_c$\\
\hline
${X}_{in}$ & 0.6594 & 0.8415 & 0.5796 & 0.7163 \\
\hline
\end{tabular}}
\label{table1}
\end{table}

Fig.\ref{Fig2} displays the TP and FP proportions when using the ER threshold and the FP when looking at the Top 10. We first notice that MKL displays less FP than SVM. In both techniques though, it is unlikely to find FP with high rankings (here arbitrarily chosen as Top 10), especially for high SNR and distributed signals.

\section{Discussion}
In this work, we investigate how SNR and sparsity affect the recovery of linear machine learning models. Our results show that for low SNR (1 or 1.5), univariate test recovery is poor while machine learning model performance, although low, can be significant. For higher SNR, all methods perform well, with MKL performing better than SVM in 20$\%$ of the cases. This confirms previous results suggesting that \emph{meaningful} grouping (here on the channels) improves performance \cite{Huang2010,Schrouff2018}. In addition, the similarity between weight maps and input difference signal is higher at the channel level than at the feature level (even when summarizing weights a posteriori), suggesting that grouping also improves recovery. All models led to false positives, suggesting that univariate tests are not immune to FP. This is an important conclusion as the univariate contrast is often referred to as the `ground truth' and used for model optimization e.g. based on weight stability \cite{Kia2017}. Regarding the machine learning models, FP are present for both machines, but only in cases of low SNR. They completely disappear in high SNR cases, and MKL seems more robust to them than SVM. In most cases, it is unlikely to encounter false positives that have a high contribution.

Overall, our results show that linear machine learning weights can, in some cases, be trusted as proxy for neural signal difference between experimental conditions. Identifying which SNR-$S$ case a specific classification/regression problem represents is however not straightforward. It seems that multiple measures would be useful, including univariate testing and model performance from both distributed and sparse models. 

As this work is based on semi-simulated data, many arbitrary choices had to be made and not all parameters have been varied (e.g. spatial correlation structure between channels). We however believe that the obtained simulated signals resemble true ECoG stimulus-based responses, on average across time points. As other modalities (fMRI, PET) are expected to display a higher spatial smoothness, performing similar simulations with other types of data would be interesting. Similarly, investigating more machine learning models would provide further insight on the influence of prior assumptions on model recovery.

\section*{Acknowledgment}

J.S. was supported by a Marie Sklodowska-Curie Actions fellowship (Project: DecoMP-ECoG, 654038). J.M.M. was supported by the Wellcome Trust (U.K.). We thank Dr. Josef Parvizi for his support and sharing of the data.

\bibliography{IEEEabrv,PRNI2018}

\end{document}